\documentclass[11pt,a4paper]{article}
\usepackage[hyperref]{eacl2021}
\usepackage{times}
\usepackage{latexsym}

\usepackage{graphicx}

\usepackage{microtype}
\usepackage{booktabs}

\aclfinalcopy 

\newcommand{\into}{$\rightarrow$}

\newcommand{\Sref}[1]{\S\ref{#1}}
\newcommand{\fref}[1]{figure~\ref{#1}}

\newcommand{\tref}[1]{table~\ref{#1}}

\title{An Exploration of Data Augmentation Techniques for Improving English to Tigrinya Translation}

\author{Lidia Kidane$^\clubsuit$ \quad Sachin Kumar$^\diamondsuit$ \quad Yulia Tsvetkov$^\diamondsuit$ \\
$^\clubsuit$African Institute for Mathematical Sciences, Kigali, Rwanda \\
$^\diamondsuit$Language Technologies Institute, Carnegie Mellon University, Pittsburgh, PA, USA \\
\texttt{\small lkidane@aimsammi.org, sachink@cs.cmu.edu, ytsvetko@cs.cmu.edu}}

\date{}

\begin{document}
\maketitle
\begin{abstract}
It has been shown that the performance of neural machine translation (NMT) drops starkly
in low-resource conditions, often requiring large amounts of auxiliary data to achieve competitive results. An effective method of generating auxiliary data is back-translation of target language sentences. In this work, we present a case study of Tigrinya where we investigate several back-translation methods to generate synthetic source sentences. We find that in low-resource conditions, back-translation by pivoting through a higher-resource language related to the target language, proves most effective resulting in substantial improvements over baselines.
\end{abstract}



\section{Introduction}

Tigrinya is a Semitic language spoken by around 8 million people in the African countries of Eritrea, accounting for more than half of its population, and Ethiopia where it is used as informal \emph{lingua franca}. However, over 60\% of the internet’s content is in English, while Tigrinya, for example, accounts for less than 0.1\% of it~\citep{w3techs}. With 40\% of the Tigrinya speakers being monolingual\footnote{\url{https://african-languages.com/tigrinya-language/}}, this essentially locks away the majority of the internet content for them. Availability of machine translation systems capable for translating English to Tigrinya and vice-versa is an imperative for its speakers to be able to function in an increasingly-global online world. 

Despite recent advances in neural machine translation~\citep{38ed090f8de94fb3b0b46b86f9133623,NIPS2017_3f5ee243}, 
such systems are difficult to develop for many African languages including Tigrinya primarily due to the lack of large amounts of high quality parallel data.
Phrase-based statistical machine translation (PBSMT)~\citep{koehn03pbsmt,DBLP:journals/corr/abs-1804-07755} has been shown to perform well under low data conditions but is challenging to develop for Tigrinya owing to its complex morphological structure~\citep{article}. 
For many low-resource languages, this challenge has led to various proposals for leveraging monolingual data that exist in either or both source and target languages, which are usually more abundant. Prior approaches include self-training~\citep{imamura-sumita-2018-nict}, transfer learning~\citep{zoph-etal-2016-transfer} and data-augmentation techniques like forward translation~\citep{zhang-zong-2016-exploiting} and back-translation~\citep{sennrich-etal-2016-improving}.

Back-translation has been used in current state-of-the-art NMT systems, outperforming other approaches in high resource
languages~\citep{ng-etal-2019-facebook} and improving performance in low resource conditions~\citep{hoang-etal-2018-iterative}. The
approach involves training a target-to-source (backward) model on the available
parallel data and using that model to generate synthetic translations of a large
number of monolingual sentences in the target language. The available authentic
parallel data is then mixed with the generated synthetic parallel data without
differentiating between the two~\citep{sennrich-etal-2016-improving} to train a final source-to-target (forward)
model. However, in low resource scenarios, the authentic parallel data available is not sufficient to train a backward model that will generate quality synthetic data. 

In this work, we explore this setting in detail. 
Combining techniques from transfer learning and back-translation, we propose several data-augmentation strategies to improve English-to-Tigrinya translation. In our experiments, we show that leveraging Amharic---a higher resource language closely related to Tigrinya---for data augmentation, gives improvements of up to +7 BLEU points over baselines. 

\section{Background and Methods}
\label{sec:methods}

We first formalize the task setup. Given a source language (\textsc{src}), a target language (\textsc{tgt}) and a typologically related language of \textsc{tgt}, \textsc{rel}, our goal is train a model $f(\cdot; \theta)$ which takes a \textsc{src} sentence $\mathbf{x}$ as input and generates its translation, $\mathbf{\hat{y}}_\textsc{tgt}=f(\mathbf{x}; \theta$). Here, $\theta$ are learnable parameters of $f$.
We are given sentence aligned \textsc{src--tgt}, \textsc{src--rel}, \textsc{rel--tgt} parallel corpora, and monolingual corpora in \textsc{rel} and \textsc{tgt}. 

In this work, we use transformer based encoder-decoder models~\citep{NIPS2017_3f5ee243} as $f$, 

We assume that the parallel \textsc{src--tgt} corpus is small, which makes training $f$ challenging~\citep{sennrich-zhang-2019-revisiting}. We now describe ways of leveraging the available monolingual data in \textsc{tgt} and resources in \textsc{rel} to generate synthetic \textsc{src\into tgt} sentences which can be augmented with the authentic \textsc{src--tgt} corpus to improve the generation quality of $f$. 

\paragraph{\textsc{BT-Direct: tgt\into src}} This is most common way to create synthetic parallel data by translating \textsc{tgt} monolingual data to \textsc{src}~\citep{sennrich-etal-2016-improving}. The backward model \textsc{tgt\into src} is trained using the available \textsc{src-tgt} parallel data. While this provides a natural way to utilize monolingual data, when the parallel data is scarce, the backward model's quality is as limited as the vanilla forward model. This results in poor quality synthetic data which is detrimental, as we show in our experiments. \citet{hoang-etal-2018-iterative} proposed an iterative BT to alleviate this issue, but this technique requires multiple rounds of retraining models in both directions which are slow and expensive. 

\paragraph{\textsc{BT-Indirect: rel\into src}} We train a \textsc{rel\into src} model using more abundant \textsc{rel-src} parallel data and use this model to translate monolingual data in \textsc{tgt} to \textsc{src}. Given that \textsc{rel} and \textsc{tgt} are closely related and written in the same script, this can serve as a proxy back-translation model allowing transfer between the two languages. 

\paragraph{\textsc{BT-Pivot: tgt\into rel\into src}} 
Despite closeness of \textsc{rel} and \textsc{tgt}, back-translating \textsc{tgt} using a \textsc{std\into src} model can result in noisy translations which can hurt the final performance of \textsc{src\into tgt}. Here, we exploit closeness of \textsc{rel} and \textsc{tgt} using the following method to generate synthetic \textsc{src--tgt} data. We train two models, one to translate \textsc{tgt\into rel} and another to translate \textsc{rel\into src}. For the former, depending on available parallel and monolingual resources in \textsc{tgt} and \textsc{rel}, the \textsc{tgt\into rel} model can be trained either in (1) a supervised manner (we refer to this setting as \textsc{BT-Pivot-Sup}), or (2) in an unsupervised manner~\citep{lample-etal-2018-phrase} (\textsc{BT-Pivot-Unsup}). The latter is trained with more easily available \textsc{src--rel} parallel data. To backtranslate a given \textsc{tgt} sentence, we first translate it to \textsc{rel} using the \textsc{tgt\into rel} model, and then to \textsc{src} using the \textsc{rel\into src} model.



\section{Experimental Setup}
\paragraph{Datasets}
We evaluate our methods with English (\textsc{en}), Tigrinya (\textsc{ti}) and Amharic (\textsc{an}) as \textsc{src}, \textsc{tgt} and \textsc{rel}.  Both \textsc{ti} and \textsc{am} are Ge’ez-scripted Semitic languages and have considerable morphological and lexical similarity~\citep{feleke-2017-similarity}.
The \textsc{en--ti} and \textsc{am--ti} parallel data are taken from Opus (JW300)~\citep{tiedemann-2012-parallel} and consist of 300K and 36K sentence pairs respectively containing text from religious domain. The \textsc{en--am} data consists of a total of 900K sentence pairs taken from Opus (JW300) and~\citet{teferra-abate-etal-2018-parallel} (News domain).
After deduplication, we created dev/test sets of 2K sentences each for both language pairs (\textsc{en-ti} and \textsc{en-am}) by randomly sampling from the JW300 corpora. We use the remaining sentences as training set.  To train unsupervised MT models, we use the \textsc{rel} and \textsc{tgt} size of the parallel corpora as the monolingual corpus. 
To create synthetic parallel data by back-translation, we create a monolingual Tigrinya corpus by crawling sentences from the official website of Eritrean Ministry of Information\footnote{\url{https://www.shabait.com/}}. After cleaning and deduplication, we get a corpus with 100K sentences.

\paragraph{Implementation and Evaluation}
We use a transformer based encoder-decoder model to conduct all our experiments~\citep{NIPS2017_3f5ee243}. We use the \textsc{base} architecture which consists of 6 encoder and decoder layers with 8 attention heads. We first tokenize all the sentences using Moses~\citep{koehn-etal-2007-moses}. For each language pair considered, we then tokenize the corpora using a BPE~\citep{sennrich-etal-2016-neural} model trained on the concatenation of the parallel corpora with 32K merge operations. We use OpenNMT-py toolkit~\citep{klein-etal-2017-opennmt} for all our experiments, with the hyperparameters recommended by~\citet{NIPS2017_3f5ee243}. We train all our supervised models (with or without data-augmentation) for 200K steps with early stopping based on validation loss. 
Finally, we evaluate the generated translations using the BLEU score~\citep{papineni-etal-2002-bleu}\footnote{While we recognize the limitations of BLEU especially for evaluating generations in morphologically rich languages~\citep{mathur-etal-2020-tangled}, METEOR~\citep{banerjee-lavie-2005-meteor} or embedding based metrics~\citep{Zhang*2020BERTScore:,sellam-etal-2020-bleurt} are simply not available for low resource languages like Tigrinya.}.

\paragraph{Baselines}
We compare the data-augmentation methods described in 
\Sref{sec:methods} 
with the following baselines

\textbf{\textsc{Unsup(src\into tgt)}} To evaluate the impact of available parallel data and feasibility of translating between unrelated languages \textsc{src} and \textsc{tgt}, we train an unsupervised NMT model to translate \textsc{src} to \textsc{tgt} using the available monolingual corpora only in the two languages.

\textbf{\textsc{Sup(src\into tgt)}} Here, we train a supervised \textsc{en\into ti} model with available parallel data only.

\textbf{Pivot through \textsc{rel}} Here, we train two translation models, a \textsc{src\into rel} model and  \textsc{rel \into tgt} model. Given a \textsc{src} test sentence, we first translate it to \textsc{rel}, which we then feed to the second model to generate text in \textsc{tgt}. We experiment with two \textsc{rel \into tgt} models leading to two baselines. First, trained with parallel supervision, we call this baseline (\textsc{Pivot:Sup(src\into rel)+Sup(rel\into tgt)};), and second, trained in an unsupervised manner, which we refer to as (\textsc{Pivot:Sup(src\into rel)+Unsup(rel \into tgt)}). 

For unsupervised \textsc{tgt$\leftrightarrow$ rel} models, we use the same architecture for this baseline and for \textsc{BT-Pivot} as described above (since it can translate in both directions). We train this model based on~\citep{lample2017unsupervised} with their recommended settings\footnote{We do not experiment with more sophisticated UNMT methods~\citep{DBLP:journals/corr/abs-1901-07291} due to their high monolingual resource requirements which are not available for either Amharic or Tigrinya.} with a few changes as follows: (1)  For each word in \textsc{tgt} vocabulary, we find its neighbors in the \textsc{rel} vocabulary within the Levenshtein distance of 2 (after removing vowel-marking diacritics). This resulted in a dictionary with 1,200 pairs. (2) We train \texttt{fasttext}~\citep{bojanowski2016enriching} embeddings for both corpora and then align them with supervision~\citep{muse} from the created dictionary. (3) We initialize the embedding tables of the encoder and decoder with the aligned embeddings and train the model parameters using autoencoding and iterative back-translation based objectives as described in~\citet{lample2017unsupervised}.

\section{Results and Analysis}

\begin{table}[]
\small
\centering
\begin{tabular}{@{}lr@{}}
\toprule
\textbf{Method} & \textbf{BLEU} \\ \midrule
\textsc{Unsup(src\into tgt)}          &  2.01             \\
\textsc{Sup(src\into tgt)}             &             7.4  \\
\textsc{Pivot:Sup(src\into rel)+Sup(rel\into tgt)}      & 8.4              \\
\textsc{Pivot:Sup(src\into rel)+UnSup(rel\into tgt)}    &5.6               \\ \midrule
\textsc{BT-Direct}             & 10.9             \\
\textsc{BT-Indirect}             & 6.2              \\
\textsc{BT-Pivot-Sup}            &11.54               \\
\textsc{BT-Pivot-Unsup}            &15.52              \\\bottomrule
\end{tabular}
\caption{BLEU scores obtained for \textsc{en-ti} translation using different baselines and data-augmentation methods described in \Sref{sec:methods}}
\label{tab:main-results}
\end{table}

The results are detailed in \tref{tab:main-results}. We observe that both unsupervised and supervised \textsc{en\into ti} models perform poorly owing to unrelatedness of English and Tigrinya and scarcity of parallel data, respectively. We get some performance improvement by first translating \textsc{en} to a related language first (Amharic in our case) and then translating it to \textsc{ti}. 

However, the gains diminish if we switch the supervised \textsc{am\into ti} model with an unsupervised one. We hypothesize this is due to small size of the monolingual corpora used to train this unsupervised model. 

Next, using a simple \textsc{ti\into en} model directly to augment data to the parallel corpus (\textsc{BT-Direct}) also gives some improvement over the best performing baseline (+2.3 BLEU). We conjecture that while additional monolingual data for Tigrinya improves the decoder language model improving the translation fluency, the improvement is hampered due to poor back-translations.

On the other hand, using a \textsc{am\into en} model for back-translating \textsc{ti} sentences results in a drop in performance likely due to very noisy examples being added to the training corpus.

We get further improvements as we consider more sophisticated methods involving pivoting through Amharic (\textsc{BT-Pivot-Sup}). We identify two potential reasons: a strong \textsc{am\into en} model on account of being trained with a larger parallel corpus, a strong \textsc{ti\into am} model owing to the similarity between two languages despite the parallel \textsc{ti \into am} model being small. 

Finally, we get the biggest jump over the best performing baseline (+7.1 BLEU) on \textsc{BT-Pivot-Unsup} where we still pivot through Amharic but \textsc{ti\into am} model trained in an unsupervised manner. We conjecture that this strong performance is due to closeness of the two languages and availability of large monolingual corpora required for training the unsupervised models.

\begin{figure}
    \centering
    \includegraphics[width=0.45\textwidth]{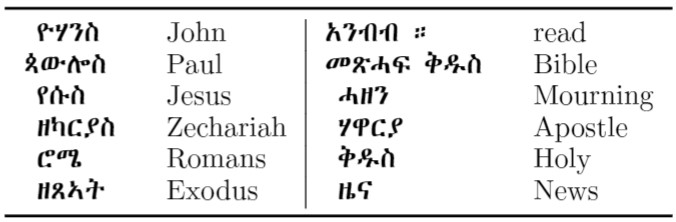}
    \caption{Examples of phrases that \textsc{BT-Pivot-Unsup} generates accurately with high frequency}
\label{tab:ngrams}
\end{figure}

\begin{figure}
    \centering
    \includegraphics[width=0.48\textwidth]{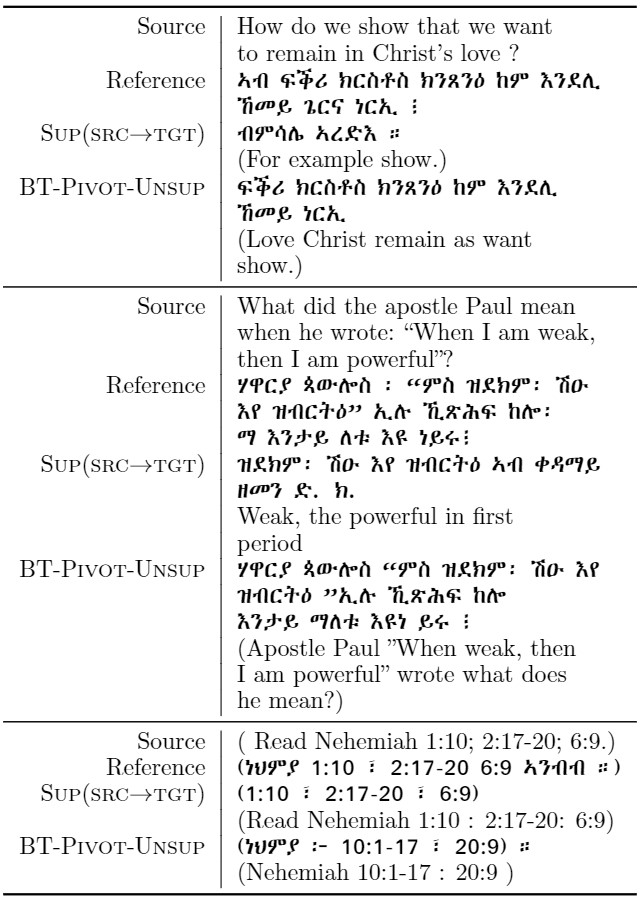}
    \caption{Selected examples of \textsc{en-ti} translations generated by \textsc{Sup(en\into ti)} and the best performing data augmentation strategy (\textsc{BT-Pivot-Unsup}).}
\label{tab:examples}
\end{figure}

To understand the influence of Amharic-based data-augmentation on the model performance, we look at the phrases in the test-set which are most-frequently generated correctly by \textsc{BT-Pivot-Unsup} using compareMT~\citep{neubig-etal-2019-compare} to extract the phrases. A small sample of such phrases is presented in \fref{tab:ngrams}. We observe that a majority of the phrases contain words shared between \textsc{am} and \textsc{ti} such as named entities or prepositions. The shared vocabulary especially benefits pivoting based back-translation where they are just copied with the \textsc{ti\into am} model resulting in their perfect translations. The \textsc{am \into en} model then is able to accurately translate it English (since it is trained on a larger parallel corpus). This is in contrast with direct \textsc{ti\into en} back-translation (\textsc{BT-Direct}) trained on a smaller parallel corpus, where these tokens often get mis-translated resulting in poorer final performance.


Finally, we present selected examples where \textsc{BT-Pivot-Unsup} performs well and compare it with examples where it suffers (compared to the baseline model \textsc{Sup(en\into tgt)} (see \fref{tab:examples}). We again observe in the examples that \textsc{BT-Pivot-Unsup} is good at generating named entities as well as tokens shared by Amharic and Tigrinya. We also note that \textsc{BT-Pivot-Unsup} fails to perform well when translating numerals (which have to be copied), often hallucinating content. We attribute these errors to noise in the back-translated data and domain mismatch between the authentic parallel corpus (containing religious text) and the synthetic parallel corpus (containing government announcements).

\section{Conclusion}
We present and compare different methods of generating synthetic parallel data and evaluate their utility for data-augmentation for low-resource machine translation. With extensive experiments on English to Tigrinya translation, we show when parallel corpora are limited, using higher-resource related languages to develop backtranslation models can lead to substantial boost in MT performance.

\section*{Acknowledgments}
This work is supported by the National Science Foundation under Grants No.~IIS2007960 and IIS2040926, and by the Google faculty research award. 

\bibliography{eacl2021}
\bibliographystyle{acl_natbib}

\end{document}